\documentclass[11pt]{article}

\usepackage{algorithm}
\usepackage{algorithmic}
\usepackage{tabularx}
\usepackage{url}
\usepackage{graphicx}
\usepackage{amsmath}

\title{On a Possible Similarity between Gene and Semantic Networks}
\author{Nicolas Turenne\thanks{nturenne.inra@yahoo.fr}}
\date{}

\begin{document}
\newtheorem{thm}{Theorem}[section]
\newtheorem{definition}[thm]{Definition}


\maketitle 

~\\ 
\begin{center}
\end{center}

\begin{abstract}
In several domains such as linguistics, molecular biology or social sciences, holistic effects are hardly well-defined by modeling with single units, but more and more studies tend to understand macro structures with the help of meaningful and useful associations in fields such as social networks, systems biology or semantic web. A stochastic multi-agent system offers both accurate theoretical framework and operational computing implementations to model large-scale associations, their dynamics and patterns extraction. We show that clustering around a target object in a set of associations of object prove some similarity in specific data and two case studies about gene-gene and term-term relationships leading to an idea of a common organizing principle of cognition with random and deterministic effects.
~

\end{abstract}









\section[Introduction]{Introduction}\label{sec:int}
Social insects can be viewed as powerful problem-solving systems with collective intelligence \cite{Deneubourg:1987},\cite{Moyson:1988},\cite{Colorni:1991},\cite{parunak:1997},\cite{bonabeau:1999}. In this paper we  describe a common framework to analyze evolution of gene networks or semantic networks and extract aggregates. 
Multiagent systems have been enriched since their origin for optimization of parallel process to nowadays framework \cite{Ferber:1999}. We used NetLogo agent-modelling toolkit \cite{Wilensky:1999} to implement automatic extraction of aggregates. NetLogo is known in agent-based community and offer a script-language, avalaible libraries to develop an \textit{ad-hoc} model. There are two tasks about this data processing. Studies about knowledge and language focuse on semantic networks by its own \cite{Fellbaum:1998} \cite{Enguehard:1993}, some studies try to associate biological constraints to phonemes formation \cite{Köhler:1986}. High-throughput studies made possible reconstruction of gene networks and enhanced gene networks modeling \cite{DeJong:2002}. But no analogy has been made between semantic network and gene network. We show in this paper that it is possible to settle a stochastic system with each relations defined explicitly and to observe how dynamics makes clusters of objects from the network represented by the set of relations.\\
Section~\ref{sec:abm} presents the stochastic multiagent framework. In Section~\ref{sec:gsn}, results show similarity between gene and semantic networks dynamics. 

\section[Agent-Based Modeling Framework]{Agent-Based Modeling Framework}\label{sec:abm}

In this section, we present basics of the stochastic multi-agent model.

\subsection[Previous studies]{Previous studies}\label{subsec:prev}

Agent-based systems become a new paradigm enabling an important step forward in empirical sciences, technology, and theory \cite{Ferber:1999}. There is no strict or commonly accepted definition of an agent. However, some common attributes can be specified for particular agent concepts. In molecular biology, agents may represent different types of enzymes acting together in a regulatory network. In linguistics, agents may represent different kinds of concepts and terms acting together into a semantic network. A reactive agent is a minimalist agent, whereas a reflexive agent would cetainly belong to a complex agent category. Between these two extreme we can find a Brownian agent approach of intermediate complexity. If interaction(s) govern(s) capabilities of an agent, an interaction in life science or cognition, can be asymetrical, where an $agent A$ is attracted by an $agent B$, but $agent B$ is repelled from $agent A$. In computational networking, the notion of an agent as a self-contained, concurrently executing software process, that encapsulates some state and is able to communicate with other agents via message passing, is seen as a natural development of the object-based concurrent programming paradigm \cite{Ferber:1990}.\\
Not all agents reacts at the same time, two possible architectures are the blackboard architecture defined by early AI community \cite{Hewitt:1973}\cite{Drogoul:1992}. It resembles a blackboard in the real-world, a data repository where agents can post and retrieve information. An alternative is the Flip-Tick architecture (FTA) based on distributd blackboards. Agents are functional units for processing data in a periodic operation called a cycle. Agents can be grouped into ensembles proceed in a synchronized way. After all agents of an ensemble have performed their cycles, a cylce counter of the ensemble representing its local time is incremented by one, as if a tick of a clock had occurred. After a tick, a new ensemble cycle occurs. Several agents may interact over many cycles by reading tags from tagborads, procesing them, and writing tags to tagboards. A tag is a data object. The communication via tagboards provides a very flexible mechanism for simulating interactions that evolve in time and/or space, as well as for parallel interactions on different spatial and temporal scales. Thus, distributed computer architectures  which are based on cooperative/competitieve ensembles of small or medium-grained agents, such as FTA, may be much more suitable for copying with time-varying interaction tasks. Some complex relations involve evolution of interactions over time, because some structures could develops over time and influence interactions requiring adaptation.\\
\cite{Schweitzer:1994}, \cite{Schweitzer:2002} and \cite{Erdmann:2003} investigates the agglomeration of active walkers on a two-dimensional surface, described by a potential $U(r,t)$, that determines the motion of walkers. Walkers are able to change $U(r, t)$ locally producing a component, that decreases $U(r, t)$ and which can diffuse and decompose. The approach is also viewed both as an agent-based framework and stochastic technique where agent moves with a random walk. Generally the technique tries to solve traditional equation of physics such as Langevin and Fokker-Planck, and spatio temoporal density distributions are analyzed across nonlinear aspects. \cite{Deutsch:1999} and \cite{Othmer:1997} emphasized interest about dynamic aggregation using random walks.

\subsection[Dynamic 2-Population Model]{Dynamic 2-Population Model}\label{subsec:dynmod}
\begin{figure}[H]
\centering
\begin{tabular}{cc}
\includegraphics[width=1.5in,height=2in]{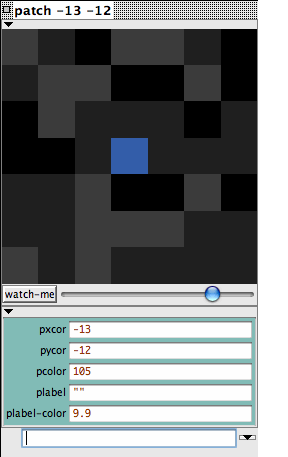} 
\end{tabular}
\caption{2-d patch.}
\label{fig:patch}
\end{figure}

Let consider a map where area is divided into elementary patches (Figure~\ref{fig:patch}).

\begin{definition} \textup{ Patch  }\hspace*{\fill} \\
\textup{ A Patch $q$ is an elementary 2-dimensional geometric area where an agent can move to. The space of movement can be settled by a vertical and horizontal number of patches. Lenght of area is defined by the amount $s$ of a patches; a specific is settle as a number of pixels. }
\end{definition} 

We need to define what is a population of objects. An object can be a gene, a linguistic term or something else.

\begin{definition} \textup{ Population  and Agent }\hspace*{\fill} \\
\textup{ A Population $P$ is defined by its size $N$ and the series of its objects \{$p_{1}$,...,$p_{N}$\}. Let consider a grid of patches \{$q_{1}$,...,$q_{s}$\}. Each object $p_{i}$ is called an agent and is assigned to a patch $q_{j}$.    }
\end{definition} 

Each agent is defined by some rules and a motility. This set of parameters constitutes the model of the multi-agent system. In our system a rule defines an interaction. Motility is a random orientation and a number of patches where agent is supposed to move.

\begin{definition} \textup{ Model }\hspace*{\fill} \\
\textup{ Let consider a set of k populations $P_{i}$ with $i \in {1,...,k}$, the total number $N$ of agent is $N=\sum_{j=1}^{j=k}\#P_{i}$. Let consider an agent $a \in P{k}$ and another agent $b \in P{m} with k \ne m$. $a$ and $b$ are defined by a same kind of motility $v_{i}$ defining a shift from current patch to another patch with a random orientation $g_{i}$; motility and orientation are states of an agent, called a Brownian agent. $a$ and $b$ are related by a link $I(a, b)$, so that $b$ modifies state of $a$.  }
\end{definition} 

If we assume that $N$ is constant:

\begin{eqnarray}
  \label{Density}
 P(r, g, t) = P(r_{1}, g_{1}, ..., r_{N}, g_{N}, t)
\end{eqnarray} 

$P(r, g, t)$ describes the probability density of finding $N$ Brownian agents with the distribution of internal parameters $g$ and positions $r$ considered. Considering both interaction (i.e. change of internal parameters $g_{i}$) and movement of the agents, probability changes over time follow the Chapman-Kolmogorov equation:

\begin{equation}
  \label{dP}
  \begin{split} 
 \frac{\partial }{\partial t} P(r, g, t) = \sum_{r'\neq r} \left\{ w(r|r')P(r', g, t) - w(r'|r)P(r, g, t)\right\} +\\
                                 \sum_{g'\neq g} \left\{ w(g|g')P(r, g', t) - w(g'|g)P(r, g, t)\right\}
                                 \end{split}
\end{equation} 

The first line of the right-hand side of Equation~\ref{dPbis} describes the change in the probability due to the motion of the agents; the second line decribes the "loss" and "gain" of agents with respect to changes in their internal parameter ($g$). The transition rates $w(r|r')$ and $w(g|g')$ refer to any possible transition within the distributions $r'$ and $g'$, respectively, which leads to the assumed distribution $r$ and $g$. In the limit of small jumps, the continuous part of Equation~\ref{dP} can be derived into discrete version and be transformed into a second-order partial equation also called Fokker-Planck equation.\\
Internal parameter $g$ can change over area (cell compartement in molecular biology, or document in electronic libraries). In first approximation we do not consider any internal parameter. So Equation~\ref{dP} becomes :

\begin{eqnarray}
  \label{dPbis}
 \frac{\partial }{\partial t} P(r, t) = \sum_{r'\neq r} \left\{ w(r|r')P(r', t) - w(r'|r)P(r, t)\right\}
\end{eqnarray} 

Discretization of space and time is revealed by polygons (i.e. patch) definifing possible orientation how agents can move. Square lattice offers 8 possible orientations, let call $d$ this number. It moves one step of constant lenght $l$ per unit of time $\Delta t$, i.e. the speed $v_{i}= l/\Delta t$ remains constant. If $r'$ is a nearest neihgbour position, then possible positions are:

\begin{eqnarray}
  \label{r'}
 r' = r + \Delta r, \Delta r = l(cos\ \tau g_{x} + sin\ \tau g_{y}), \tau = \frac{n \pi}{3}, n = {1,...,8}
\end{eqnarray} 

For the transition rate with a jump of one per unit of time and a non-biased walker is :

\begin{eqnarray}
  \label{w}
 w(r'|r) = \frac{1}{d}, w(r|r) = 0
\end{eqnarray} 

For normalisation $\sum_{d}w(r'|r) = w = 1$. \\
For a Brownian agent acting as a random walker with interaction, a potential $h(r,t)$ on a landscape can symbolize an external force (i.e. interactions). The potential field influences movement through the gradient of the field, so transitions changes. Using the gradient :

\begin{eqnarray}
  \label{dh}
 \frac{ \partial h(r, t)}{\partial r} = \frac{   h(r + \Delta r, t) - h(r - \Delta r, t) }{ |(r + \Delta r) - (r - \Delta r) |}.
\end{eqnarray} 

Transition to $r'$ becomes with addition of a field and stochastic effects:

\begin{eqnarray}
  \label{wr}
 w(r + \Delta r |r) = \frac{1}{d} \left\lbrack 1 + \beta \frac{   h(r + \Delta r, t) - h(r - \Delta r, t) }{ |2\ \Delta r|} \right\rbrack.
\end{eqnarray} 
$\beta$ acts as a dimensional constant. For a vanishing gradient $h(r + \Delta r, t) - h(r - \Delta r, t)$ and the random walker changes its position with equal probabilities for the possible directions.

Finally the master equation Equation~\ref{dP} becomes for one agent and its associated density $p(r, g, t)$:

\begin{eqnarray}
  \label{dpFP}
 \frac{\partial }{\partial t} p(r, g, t) = - \frac{\partial}{\partial r} a_{1}(r)\ p(r, g, t) + \frac{1}{2} \frac{\partial ^{2}}{ \partial r^{2}} a_{2}(r)
\end{eqnarray} 

With :
\begin{eqnarray}
  \label{a1a2}
  \left\{
   \begin{array}{l}
 a_{1}(r) = \sum_{g'} (r' - r)\ w(r'|r),\\
 a_{2}(r) = \sum_{g}  (r' - r)^{2}\ w(r'|r).
 \end{array}
  \right.
\end{eqnarray} 

or 
\begin{eqnarray}
  \label{a1a2bis}
  \left\{
   \begin{array}{l}
  a_{1}(r) = w\ l\ \beta\ \frac{ \displaystyle \partial h(r, t)}{\displaystyle \partial t},\\
  a_{2}(r) = w\ l^{2}.
  \end{array}
  \right.
\end{eqnarray} 

We could derive such result with Langevin equation superposing deterministic and stochastic effects on Brownian particles. An adaptative landscape serves as an interaction medium for agents. This idea has similarity to communication via tagboards. Data read and written are considered as a structural information which is stored in a blackboard. Hence communication among agents can be seen as indirect communication. In a traditional stochastic multiagent system, field can be a quantity present around a given agent, and leave out by another agent in its neighbourhood. In our case the field is related by presence/absence of another agent in its close neighbourhood. So :

\begin{eqnarray}
  \label{Pot}
 h(r_{i} + \Delta r, t) = \sum_{j = 1, i \neq j}^{k} \delta( (r_{i} + \Delta r) - r_{j} ) I(i, j)
\end{eqnarray} 

It means that potential $h(r', t)$ for a given agent $a_{i}$ is the sum of interactions $I(i, j)$ that takes values 1 if an agent $a_{j}$ from population $j$ is a neighbour of agent $a_{i}$. Hence this interaction contributes to make agent to move from patch $q( a(i) )$ to a patch $q( a(j) )$. ~\\

We may defined some indices of homogeneity. First index is a critical density $d_{c}$ for which all agents are equidistributed on the landscape and there free movement is frozen because of very near interactions. $N_{p}$ is the number of patches on the landscape.

\begin{eqnarray}
  \label{Np}
 N_{p} = A^{2}
\end{eqnarray} 
Where A is the size of the landscape (in unit of patches). Hence  critical density is:

\begin{eqnarray}
  \label{dc}
 d_{c} = \frac{ n - \frac{ 3} {4}.n} { N_{p} }
\end{eqnarray} 

Where $n$ is the number of agents of a given population, and $a$ is size of an agent in term of a percent of the patch size. A patch having a size of 1 (area unit). For a high amount of agents, $\frac{ 3 } { 4 }$  is the amount of agents we lost as a direct nearest neighbours which will not move away but directly jump to a neighbour, in this way stochastic effect vanishes. Let suppose we lay out 8 agents as equidistribution on a ladscape of 60 patches and separated at least by one patch each other, 45 patches are forbidden for layout to keep freedom of agents, hence 75\% of patches.

Then for $d_{c} = 1$, critical number of agents $n_{c}$ should be:

\begin{eqnarray}
  \label{nc}
 n_{c} = \frac{  N_{p} } { 4 } 
\end{eqnarray} 

A second indice is the equidistribution of all population $P_{k}|\ k\in\{1,...,N\}$ on the landscape leading to a complete frozen movement of agents if they have interaction because they can stay at their initial patch with other agents to which they interact with. For a given population $P_{i}$ we define set $S_{i}$ of its neighbourhood around a distance $d$ (in pixel unit).

\begin{eqnarray}
  \label{Si}
 S_{i} = \left\{ P_{k} |\ \exists a_{j},\ a_{j} \in P_{k}\ and\ | r_{i} - r{j} | \leq d \right\}
\end{eqnarray} 

In the case of equidistribution of all populations, we get $S_{i}^{eq}$:
\begin{eqnarray}
  \label{Sieq}
 S_{i}^{eq} = \left\{ P_{k} |\ k \in \{1,..., N\}\ \right\}
\end{eqnarray} 

\subsection[Toy example]{Toy example}\label{subsec:toy}

\begin{figure}[H]
\centering
\begin{tabular}{cc}
\includegraphics[width=2in,height=2.5in]{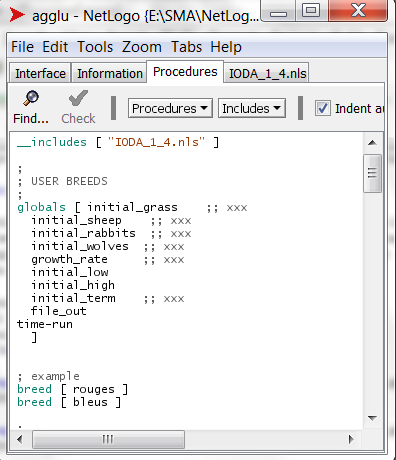} 
\end{tabular}
\caption{Netlogo language.}
\label{fig:language}
\end{figure}

We used NetLogo tool to implement our model \cite{Wilensky:1999}. Lots of others tools are available to achieve multi agent systems \cite{Nikolai:2009}. Few of them propose a swarming modeling framework. Figure~\ref{fig:language}) shows an example of NetLogo agent language, in this case for population declaration and global variable; it is formerly a declarative programming language.

Interactions are specified by a matrix specification (Figure~\ref{fig:matrix}) in a procedural way \cite{Kubera:2008}. It is quite practical to define
an interaction as a binary relation between two populations, set by an operator (Figure~\ref{fig:interaction}). Each operator is defined by primitives of Netlogo language. Interactions and matrix are written in separate files. A second interesting point is that no limit bounds interaction definition and in some context there is not only global interaction and some types of agents such as (electron, proton, neutron, photon) in electromagnetism, or (big mass, small mass, vacuum) in gravitation. In molecular biology, social systems, or semantic analysis, several ten thousands different populations of agent may exist and interact, and so may be specified in an artificial system. So distinct definitions of specific interactions and relations between populations is of great interest.  

\begin{figure}[H]
\centering
\renewcommand{\arraystretch}{1.2}
{\scriptsize
\begin{tabularx}{\textwidth\tt}{|X|}
\hline
interaction walk\\
actions random-walk deactivate-none\\
end\\
\\
interaction cooc\\
actions follow-path deactivate-source\\
end\\
\hline
\end{tabularx} 
}
\caption{ Interaction rules, here definition of \texttt{walk} and \texttt{cooc}. }
\label{fig:interaction}
\end{figure}

\begin{figure}[H]
\centering
\renewcommand{\arraystretch}{1.2}
{\scriptsize
\begin{tabularx}{\textwidth\tt}{|X|}
\hline
\textit{; source-family interaction-name priority cardinality $<$target-family distance$>$}\\
particles walk 0 0\\
walkers   walk 0 0\\
particles cooc 1 1 walkers 2\\
\hline
\end{tabularx} }
\caption{ Matrix of interactions, here definition of interaction between $particles$ and $walkers$ populations. }
\label{fig:matrix}
\end{figure}

\begin{figure}[H]
\centering{
\begin{tabular}{ccc}
t = 0&t = 3&t = 10\\
(a) \includegraphics[width=1.0in,height=1.0in]{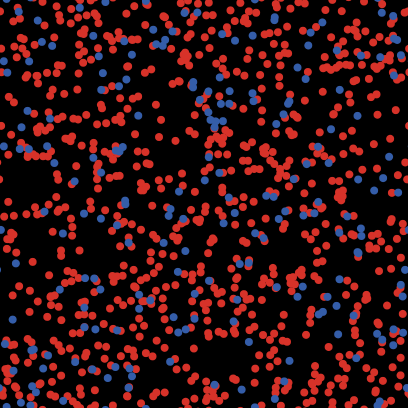}&\includegraphics[width=1.0in,height=1.0in]{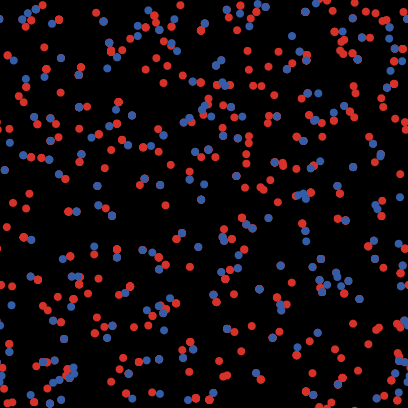}&\includegraphics[width=1.0in,height=1.0in]{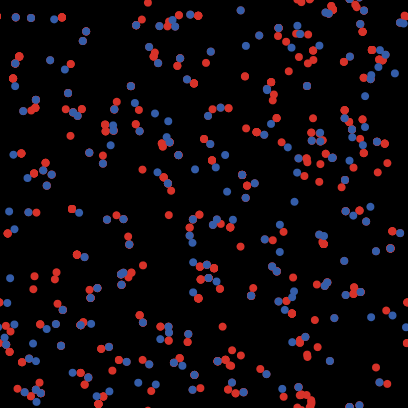}\\
(b) \includegraphics[width=1.0in,height=1.0in]{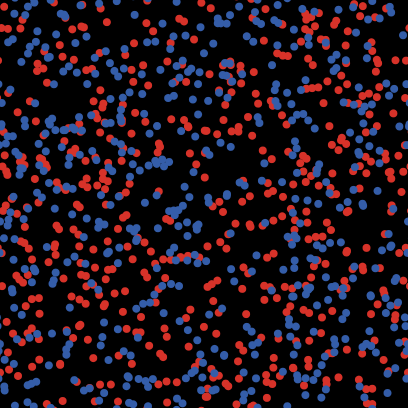}&\includegraphics[width=1.0in,height=1.0in]{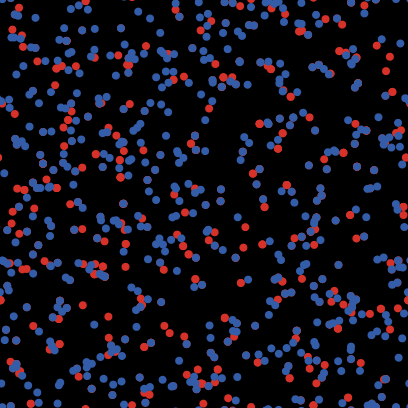}&\includegraphics[width=1.0in,height=1.0in]{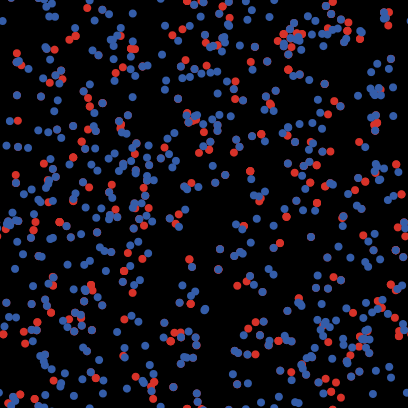}\\
(c) \includegraphics[width=1.0in,height=1.0in]{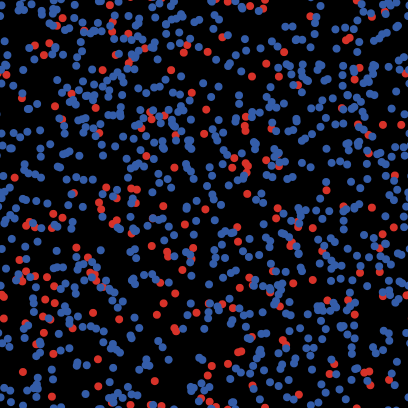}&\includegraphics[width=1.0in,height=1.0in]{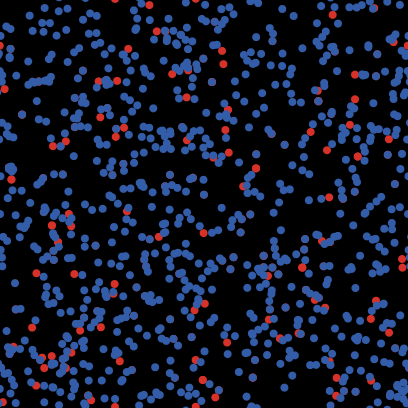}&\includegraphics[width=1.0in,height=1.0in]{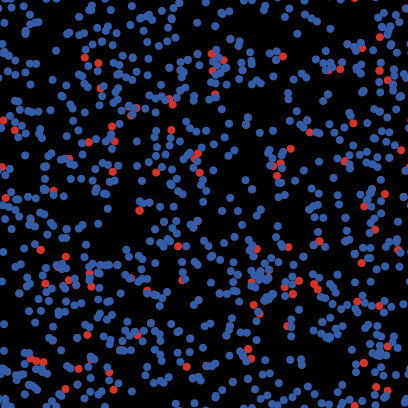}\end{tabular}
}
\caption{ 200 blue 800 red  (a, up), 500 blue 500 red  (b, centre), 800 blue 200 red (c, bottom)}
\label{fig:displaytoy}
\end{figure}

The agents, called turtles in NetLogo, represent objects. They have physical properties such as localization (patch), color, form. \\
Figure~\ref{fig:displaytoy} shows a result obtain with the model presented in previous section settled with two populations (red and blue circles).
As shown by previous figures about interaction population $walkers$ is blue, population $particles$ is in red. In practice interaction has been defined so that a $walkers$ go to neighbourhood of $particles$ agent. Figure~\ref{fig:displaytoy} shows three configurations of population size (up: 200 $particles$ and 800 $walkers$, centre: 500 $walkers$ and $particles$, bottom: 800 $walkers$ and 200 $particles$), and three configurations of time duration (t = 0, t = 3 and t = 10). Results show that independantly of size population, red population aggregates to blue population very fast (t = 3), and some fluctuation persists among red agents in any case at a long time (t = 10).

As $A = 51$, $N_{p} = 2,601$ patches, $n_{c} = 650$. In this toy example, the configuration (c) (Figure~\ref{fig:displaytoy}) having population under the critical density shows a result enough as good as other configuration.

\section[Genetic and semantic networks]{Genetic and semantic networks}\label{sec:gsn}

In this section, we present large-scale case studies.

\subsection[Data collections]{Data collections}\label{subsec:datac}

\textbf{Gene lists}

First data collection is associated to a specific molecular biological issue about trophoblast development. The trophoblast is an embryonic epithelium which combines three main processes among which proliferation and differenciation. Specificity of the bovine trophoblast is its exponential growth, which does not exist in rodents or primates. conversely, the ruminant trophoblast does not seem to invade the uterus, when the trophoblast from rodents and primates do. The question we ask here is to mine the literature concerning proliferation processes and differenciation processes. The first one is well documented in human cancer cells, the second is well studied in mouse embryos. Our aim was thus to study bovine, human and murine datasets for which respectively we get microarrays time series linked to bovine trophoblast, human cancer and mouse stem cells, even if the time series were different. In bovine species 1,975 genes are involed in this process. To solve the problem we try to extract relatioship across species between each pair of genes in a collection of documents, called a corpus. It contains 29,000 documents. The gene names had been used to make a query and retireve these document (except some of ones having big polysemy (\textit{set}, \textit{via}, \textit{p}). The number of genes included in the corpus is 655 and between them we had extracted 2,752 uniques relationships based on cooccurrence in the corpus.\\
A second data collection is associated to a specific molecule: TOR. Target Of Rapamicyn - is a molecule playing a key role in cell development and its metabolism. TOR is present in all living organisms but interactome seems to vary from a species to another (i.e. sub-interactomes). About Arabidopsis Thaliana species, less knowledge are available. Knowledge extraction of a global interactome from scientific literature across species could ensure hypotheses concerning TOR sub-interactome in A.Thaliana. At least 53 components (certainly more) are known to regulate TOR or be regulated by TOR. A corpus of 14,000 documents about Tor has been defined.

~\\
\textbf{Terminological lists}

A corpus is the dataset and consists of 1,280 abstracts in the field of material chemistry. Some experts had defined 6,603 terms of the domain and their anchors in the documents with xml tags. Corpus has been defined as a reference for term recognition \cite{Enguehard:2003}. This is a sample of tagging:

\begin{tabbing}
\small $<$notice\ id$=$"1"/$>$\\
 \hspace*{1cm} \= \small $<$variante\ refterme$=$"5564"\ statut$=$"novar"\ debut$=$"12"\ fin$=$"13"/$>$\ \\
 \hspace*{1cm} \small $<$/variante$>$ \\
 \hspace*{1cm} \= \small $<$texte$>$Xi\ is\ the\ dimensionless\ correlation\ length\ of\ the\ pair\ \\ 
 \hspace*{2cm} \= \small $<$ancre\ id$=$"12"\/$>$correlation\ function $<$ancre\ id$=$"13"/$>$.$<$/texte$>$\\
\small $<$/notice$>$
\end{tabbing}

In the previous example, abstract number 1 has been tagged with only one occurrence identified about term "correlation function". It occurs between tags $<$ancre id="12"\/$>$ and $<$ancre id="13"\/$>$. Identity of term is : 5,564. By cooccurrence analysis, we have found 9,591 relationships between pairs of terms occurring once (or more) in a same context.

\subsection[Run configuration]{Run configuration}\label{subsec:run}

\begin{figure}[H]
\centering{
\begin{tabular}{cccc}
t = 0&t = 40&t = 150&t = 1000\\
\includegraphics[width=1.0in,height=1.0in]{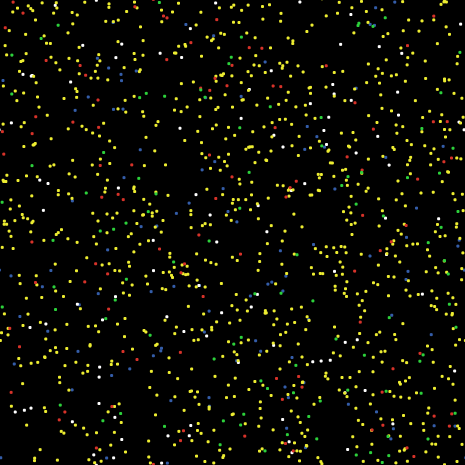}&\includegraphics[width=1.0in,height=1.0in]{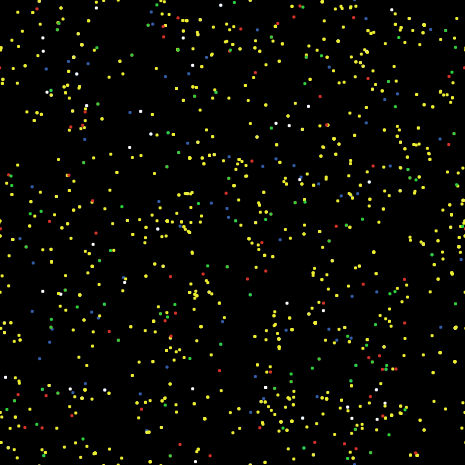}&\includegraphics[width=1.0in,height=1.0in]{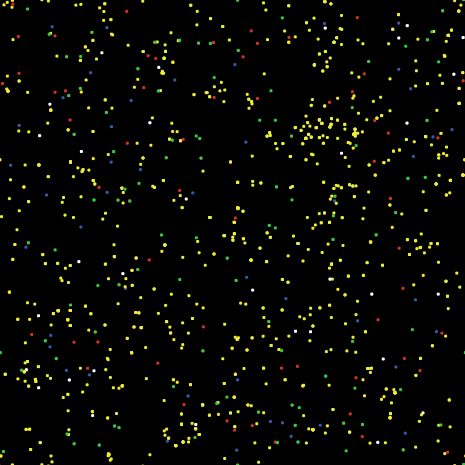}&\includegraphics[width=1.0in,height=1.0in]{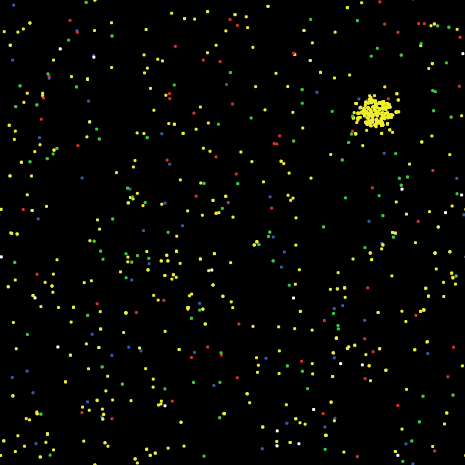}
\end{tabular}
}
\caption{ Display about 12 populations. }
\label{fig:SmallSets}
\end{figure}

Interactions operators are the same defined as seen in Section~\ref{subsec:toy}: \texttt{walk} and \texttt{cooc}. Section~\ref{subsec:sms} shows a small set with more than only two populations. It involves ten populations. Section~\ref{subsec:rws} reveals more large-scale relations, ranged from 37 to 515 populations, and 38 to 9,234 relations. These relations are extracted from validated facts in technical literature.\\
Display is made by $A = 31$ patches inline so $N_{p} = 961$ patches, and critical density is $n_{c} = 240$ agents. In experiments population size is 100. Figure~\ref{fig:SmallSets} shows results at different timepoints from $t = 0$ to $t = 1000$.  

\begin{figure}[H]
\centering{
\includegraphics[width=1.5in,height=1.5in]{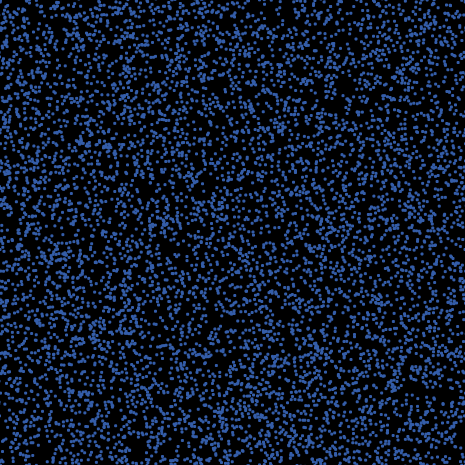}\\
}
\caption{ Display about terminological populations. }
\label{fig:BigSetInit}
\end{figure}

Figure~\ref{fig:BigSetInit} shows init display with 515 populations from the terminological set.

\subsection[Small set]{Small set}\label{subsec:sms}

The small set experiment, presented hereafter, do not gather all possible and real-world relations between components of an agent set. But contrary to the toy experiment in Section~\ref{subsec:toy}, complexity increases and becomes more realistic because of cross-relationships, as $agent\ A$ interacts with $agent\ B$, $agent\ B$ interacts with $agent\ C$ and $agent\ C$ interacts with $agent A$.\\
So question is: does agent dynamics converge to an interpretable neighbourhood ?\\
Figure~\ref{fig:matrixsmall} shows the sets of relations between twelve populations. In this artificial experiment, such one in Section~\ref{subsec:toy}, our target population is $walkers$ and the relations associated to this population is $particles$ and\\
$ab\_initio\_calculations$. Eleven other relationships are also defined to generate noise in the system. And all populations make a random walk.

\begin{figure}[H]
\centering
\renewcommand{\arraystretch}{1.2}
{\scriptsize
\begin{tabularx}{\textwidth\tt}{|X|}
\hline
\textit{; source-family interaction-name priority cardinality $<$target-family distance$>$}\\
particles walk 0 0\\
walkers walk 0 0\\
ab\_initio\_calculations walk 0 0\\
abductor\_digiti\_minimi walk 0 0\\
abductor\_pollicis\_brevis walk 0 0\\
aberrant\_activation walk 0 0\\
aberrant\_methylation walk 0 0\\
aberrant\_regulation walk 0 0\\
abnormal\_magnetic walk 0 0\\
abnormal\_representation walk 0 0\\
absolute\_expression walk 0 0\\
abundance\_proteins walk 0 0\\
abundant\_transcripts walk 0 0\\
particles cooc 1 1 walkers 2\\
ab\_initio\_calculations cooc 1 1 walkers 2\\
ab\_initio\_calculations cooc 1 1 abductor\_digiti\_minimi 2\\
abductor\_digiti\_minimi cooc 1 1 abductor\_pollicis\_brevis 2\\
abductor\_pollicis\_brevis cooc 1 1 aberrant\_activation 2\\
aberrant\_activation cooc 1 1 aberrant\_methylation 2\\
aberrant\_methylation cooc 1 1 aberrant\_regulation 2\\
aberrant\_regulation cooc 1 1 abnormal\_magnetic 2\\
abnormal\_magnetic cooc 1 1 abnormal\_representation 2\\
abnormal\_representation cooc 1 1 absolute\_expression 2\\
absolute\_expression cooc 1 1 abundance\_proteins 2\\
abundance\_proteins cooc 1 1 abundant\_transcripts 2\\
abundant\_transcripts cooc 1 1 abductor\_digiti\_minimi 2\\
\hline
\end{tabularx} }
\caption{ Matrix of interactions. }
\label{fig:matrixsmall}
\end{figure}

Result, shown on Figure~\ref{fig:SiSmall}, is different over time. At beginning all populations are equidistributed around $walkers$ agents but after 20 timepoints $ab\ initio\ calculations$ and $particles$, are more present than other populations. It seems consistant with the kind of interactions defined with $walkers$ population. But globally all populations surround $walkers$ agents in average. In fact, frequency of the two close populations of $walkers$ population is twice more than the average agent frequency in the neighbourhood.

\begin{figure}[H]
\centering
\renewcommand{\arraystretch}{1.2}
{\scriptsize
\begin{tabularx}{12cm}{|>{\centering\hsize=0.8\hsize\arraybackslash}X|>{\centering\hsize=0.4\hsize\arraybackslash}X|>{\centering\hsize=0.4\hsize\arraybackslash}X|>{\centering\hsize=0.4\hsize\arraybackslash}X|}
   \hline
    \textbf{population label}   &  \textbf{$t = 0$}   &   \textbf{$t = 20$}   &   \textbf{$t = 1000$}    \\
   \hline
	\textbf{ab\_initio\_calculations}		&	\textbf{21} & \textbf{51} & \textbf{45} \\
	abductor\_digiti\_minimi							& 23 				& 31 					& 19 \\
	abductor\_pollicis\_brevis						& 29 				& 29 					& 28 \\
	aberrant\_activation									& 31 				& 26 					& 27 \\
	aberrant\_methylation									& 26 				& 30 					& 21 \\
	aberrant\_regulation									& 25 				& 43 					& 17 \\
	abnormal\_magnetic										& 20 				& 41 					& 16 \\
	abnormal\_representation							& 23 				& 35 					& 15 \\
	absolute\_expression									& 27 				& 29 					& 16 \\
	abundance\_proteins										& 23 				& 33 					& 17 \\
	abundant\_transcripts									& 31 				& 32 					& 16 \\
	\textbf{particles	}									& \textbf{23} & \textbf{61} & \textbf{55} \\
   \hline
  $<average>$														& 27 				& 34 					& 21 \\
   \hline
\end{tabularx} } 
\caption{ $S_{walkers}$ , set of neighbours of $walkers$ population at different timepoints. }
\label{fig:SiSmall}
\end{figure}

\subsection[Real-world sets]{Real-world sets}\label{subsec:rws}

Figure~\ref{fig:targets} shows parameters with three target populations : $actb$ about embryo corpus, $adsorption$ about material chemistry corpus, and $tor$ about TOR molecule corpus. One achieved two experiments with each target population each one corresponding to two different initialization sets of relations. The first set is called "restricted" and is related to only relations of a target population with all populations with which it is linked. The other kind of experiment, called "extented", gathers all populations co-occuring with the target population, and populations linked to these co-occuring populations; the relations settled as init parameters are relations between these all populations.

\begin{figure}[H]
\centering
\renewcommand{\arraystretch}{1.2}
{\scriptsize
\begin{tabularx}{7.7cm\tt}{|l|r|r|}
   \hline
    \textbf{target agent label}   &  \textbf{\#relations}   &  \textbf{\#populations}    \\
   \hline
	actb (extended)							& 1511 				& 474 		 \\
	actb												& 38 				  & 37 		 \\
	adsorption (extended)				& 9234 			 	& 515 		 \\
	adsorption								  & 77 				  & 76 		 \\
	tor (extended)							& 3728 				& 416 		 \\
	tor												  & 397				  & 396 		\\
   \hline
\end{tabularx} } 
\caption{ Target population label and the interaction/relation context. }
\label{fig:targets}
\end{figure}

We had extracted genes interactors from several databases :
\begin{itemize}
	\scriptsize \item http://www.genecards.org
	\scriptsize \item http://www.signaling-gateway.org
  \scriptsize \item http://www.compbio.dundee.ac.uk/www-pips
	\scriptsize \item http://www.wikigenes.org
	\scriptsize \item http://en.wikipedia.org/
	\scriptsize \item http://biomyn.de
	\scriptsize \item http://biit.cs.ut.ee/graphweb/exampleInput/Human\_protein\_interactions\_\%28IntAct\%29.txt
	\scriptsize \item http://www.cbs.dtu.dk/courses/27619/humanPPint.lst
	\scriptsize \item http://www.ncbi.nlm.nih.gov/IEB/Research/Acembly/
	\scriptsize \item http://www.ihop-net.org/UniPub/iHOP/
	\scriptsize \item http://www.ebi.ac.uk/intact/
\end{itemize}

~\\
\textbf{Experiments on gene sets - focus on "Tor"}

Rapamycin is a potent antibiotic produced by a strain of \textit{Streptomyces Hygroscopicus} isolated from a soil sample collected in Rapa-Nui (Easter Island). The pharmaceutical potential of rapamycin was originally discovered in a screen for novel antifungal agents. Much later, rapamycin was found to exhibit immunosuppressive activity due to its capacity to block the growth and proliferation of T cells. More recently, rapamycin has been found to display anticancer properties. The atypical Ser/Thr kinase TOR (Target of Rapamycin) is a central controller of cell growth that is structurally and functionally conserved in all species. Four major inputs control mammalian TOR (mTOR): nutrients, such as amino acids; growth factors, such as insulin; cellular energy levels, such as the AMP:ATP ratio; and stress, such as hypoxia. mTOR controls cell growth by the positive and negative regulation of several anabolic and catabolic processes, respectively, that collectively determine cell size. These cellular processes include translation, ribosome biogenesis, nutrient transport, autophagy, and AGC kinase activation.\\
Figure~\ref{fig:TorSets} shows neighbourhood result of $tor$, in our multiagent modeling; the first 20 most frequent population of $S_{tor}$ for restricted and extended relations. Populations are different. Only population $cox2$ is common.\\

\begin{figure}[H]
\centering{
 \scriptsize
\begin{tabular}{cc}
restricted&extended\\
\begin{tabularx}{6cm\tt}{|l|X|}
   \hline
    \textbf{population label}   &  Neighbour Frequency    \\
   \hline
sec14		&	6\\
ccc1		& 5\\
cdc42		& 5\\
lat1		& 5\\
sap185	&	5\\
sip2		& 5\\
big1		& 4\\
cdc10		& 4\\
cox2		& 4\\
cyt1		& 4\\
dal4		& 4\\
eap1		& 4\\
fap1		& 4\\
fhl1		& 4\\
fol2		& 4\\
grr1		& 4\\
gtr1		& 4\\
mlh1		& 4\\
pim1		& 4\\
pph21		& 4\\
   \hline
$<global\ average>$	& 2.0				 \\
   \hline
\end{tabularx} 
 
&
\begin{tabularx}{6cm\tt}{|l|X|}
   \hline
    \textbf{population label}   &  Neighbour Frequency    \\
   \hline
bub2	&	12\\
rce1	&	9\\
slh22	&	9\\
pkc1	&	8\\
sec1	&	8\\
apg7	&	7\\
bck2	&	7\\
cox2	&	7\\
mak31	&	7\\
mgm1	&	7\\
plc1	&	7\\
prs1	&	7\\
sec18	&	7\\
tad2	&	7\\
cdc37	&	6\\
hap4	&	6\\
hmo1	&	6\\
maf1	&	6\\
mak3	&	6\\
pfk1	&	6\\
   \hline
$<global\ average>$	& 2.6				 \\
   \hline
\end{tabularx} 
 
\end{tabular}
}
\caption{ Display about $tor$ population : restricted (left) and extended (right). }
\label{fig:TorSets}
\end{figure}

From databases we have identified 13 ortholog names, i.e. identical gene with different names in different species. And 49 variants names. It helped to extract candidate interactors stored in public databases, 414 has been found in possible interaction with TOR in any species. Running our model we find 45 populations having frequency over 2 times the average. After comparison with the set of interactor gene names we find 11 genes in common : $cdc25$, $map2$, $pho80$, $pkc1$, $plc1$, $rrn9$, $rsc9$, $sfp1$, $sik1$, $snf3$, $spr2$.

~\\
\textbf{Experiments on gene sets - focus on "ActB" }

Beta actin is one of six different actin isoforms which have been identified. ACTB is one of the two nonmuscle cytoskeletal actins. Actins are highly conserved proteins that are involved in cell motility, structure and integrity. Actin is a ubiquitous protein involved in the formation of filaments that are major components of the cytoskeleton evolving in the cell cytoplasm.
Figure~\ref{fig:ActbSets} shows neighbourhood result of $actb$, in our multiagent modeling; the first 20 most frequent population of $S_{actb}$ for restricted and extended relations. Populations are quite well different. Populations in common  are : $arf1$ , $ctsb$ , $fau$ , $gnb2l1$ , $nebl$ , $ppia$ , $prpf8$ , $rps23$ , $tpi1$.

\begin{figure}[H]
\centering{
 \scriptsize
\begin{tabular}{cc}
restricted&extended\\
\begin{tabularx}{6cm\tt}{|l|X|}
   \hline
    \textbf{population label}   &  Neighbour Frequency    \\
   \hline
arf1		&	7\\
gnb2l1	&	7\\
eef1a1	&	6\\
hspa5		&	6\\
ctsb		&	5\\
fau			&	5\\
gapdh		&	5\\
jun			&	5\\
myc			&	5\\
pou5f1	&	5\\
rps23		&	5\\
sparc		&	5\\
tpi1		&	5\\
ubc			&	5\\
des			&	4\\
nebl		&	4\\
pgk1		&	4\\
ppia		&	4\\
prpf8		&	4\\
rpl17		&	4\\
   \hline
$<global\ average>$	& 3.9				 \\
   \hline
\end{tabularx} 
&
\begin{tabularx}{6cm\tt}{|l|X|}
   \hline
    \textbf{population label}   &  Neighbour Frequency    \\
   \hline
prpf8		&	22\\
nebl		&	19\\
casc3		&	13\\
ctsb		&	11\\
fau			&	11\\
rplp0		&	11\\
rpl41		&	9\\
uba52		&	9\\
gnb2l1	&	8\\
rps23		&	8\\
tpi1		&	8\\
anxa2		&	7\\
arf1		&	7\\
rps4		&	6\\
ccni		&	5\\
chd4		&	5\\
clp1		&	5\\
hspa9		&	5\\
mybbp1a	&	5\\
ppia		&	5\\
   \hline
$<global\ average>$	& 2.2 				 \\
   \hline
\end{tabularx} 
 
\end{tabular}
}
\caption{ Display about $actb$ population : restricted (left) and extended (right). }
\label{fig:ActbSets}
\end{figure}

From databases we have identified none ortholog names. And 2 variants names. We extracted 299 candidates interactors stored in public databases and in any species. Running our model we find 37 populations having frequency over 2 times the average frequency. After comparison with the set of interactor gene names we find 11 genes in common : $anxa1$, $anxa2$, $arf1$, $gnb2l1$, $hspa9$, $nebl$, $rpl17$, $rpl4$, $rplp0$, $rps23$, $rps4$.

~\\
\textbf{Experiments on terminological sets - focus on "adsorption"}

Adsorption is the process of attraction of atoms or molecules from an adjacent gas or liquid to an exposed solid surface. Such attraction forces (adhesion or cohesion) align the molecules into layers ("films") onto the existent surface.\\
Figure~\ref{fig:AdsSets} shows neighbourhood result of $adsorption$, in our multiagent modeling; the first 20 most frequent population of $S_{adsorption}$ for restricted  and extended relations. Populations are quite well different. Populations in common  are : $ammonia$ , $benzene$ , $coadsorption$ , $electrode\ potential$ .

\begin{figure}[H]
\centering{
 \scriptsize
\begin{tabular}{cc}
restricted&extended\\
\begin{tabularx}{6.2cm\tt}{|l|X|}
   \hline
    \textbf{population label}   &  Neighbour Frequency    \\
   \hline
microprobe											&	7\\
\_l															&	5\\
coadsorption										&	5\\
dissociation										&	5\\
electrode\_potential						&	5\\
ellipsometry										&	5\\
reflection											&	5\\
review													&	5\\
alcohol													&	4\\
ammonia													&	4\\
diagram													&	4\\
electron\_microprobe\_analysis	&	4\\
heat\_of\_adsorption						&	4\\
mixture													&	4\\
motion													&	4\\
pyridine												&	4\\
synchrotron\_radiation					&	4\\
\_m															&	3\\
band\_structure									&	3\\
benzene													&	3\\
   \hline
$<global\ average>$							& 2.5 				 \\
   \hline
\end{tabularx}  
&
\begin{tabularx}{5.5cm\tt}{|l|X|}
   \hline
    \textbf{population label}   &  Neighbour Frequency    \\
   \hline
acetic\_acid						&	18\\
growth\_mechanism				&	16\\
reactivity							&	15\\
palladium\_complex			&	14\\
electrode\_potential		&	12\\
trend										&	10\\
molecular\_beam					&	9\\
molecular\_mass					&	9\\
ammonia									&	8\\
carbon\_monoxide				&	8\\
coadsorption						&	8\\
copper\_chloride				&	8\\
electromagnetic\_field	&	8\\
substructure						&	8\\
austempering						&	7\\
physisorption						&	7\\
sandwich\_structure			&	7\\
secondary\_martensite		&	7\\
benzene									&	6\\
   \hline
$<global\ average>$			& 2.6 				 \\
   \hline
\end{tabularx}  
 
\end{tabular}
}
\caption{ Display about $adsorption$ population : restricted (left) and extended (right). }
\label{fig:AdsSets}
\end{figure}

Running our model we find 49 populations having frequency over 2 times the average frequency. We extracted 1006 singles words stored in Wikipedia page in English after removing stop words. After comparison with the set of Wikipedia words we find 11 phrases in common : $acetic\_acid$, $growth\_mechanism$, $electrode\_potential$, $molecular\_beam$, $molecular\_mass$, $carbon\_monoxide$, $copper\_chloride$, $electromagnetic\_field$, $physisorption$, $sandwich\_structure$, $binding\_energy$, $electron\_microprobe\_analysis$, \\
$heat\_of\_adsorption$, $phase\_diagram$, $skin\_effect$, $surface\_energy$, \\
$surface\_reaction$, $surface\_temperature$, $electron\_energy\_loss$. It means that one single word in these phrases occurs in the Wikipedia page. If we match only whole phrases in common :  $acetic\_acid$,  $carbon\_monoxide$ and $physisorption$ occur in the Wikipedia page.

\section[Conclusion]{Conclusion}
A Brownian agent model defines a strong theoretical framework mixing stochastic and deterministic aspects of self-organization over time. Time, order and stochasticity are main parameters of real-world natural systems in which complexity is due to large number of components and can not be explained by traditional function modeling. This theoretical framework is also implicitly the core of a pragmatic implementation as a large-scale multi-agent system ensuring observation of dynamics and convergence of a specific required relational model. Model is itself defined by a set of rules and populations of agents. We adopt this framework to propose an aggregative model of gene sets dynamics (genetic net) and term set dynamics (semantic net). The set of rules are defined through a set of interactions primitives and a matrix of relations between populations. The knowledge of relations comes from large-scale literature mining. We show that choosing a target population, its neighbourhood is not random and consistant to reality either with a gene neighbourhood and a semantic neighbourhood, with a same way of self-organization (i.e. contextual aggregation). It should lead us to think in part that the way of agregation between terms could be inspired from the way of genes organization. In this way, semantic nature of language is not fully deterministic but contains also a stochastic part not guided by intrinsic organization of biology. Such principle should not be so surprising since any language speaker is, initially, a biological entity.

\section[Acknowledgments]{Acknowledgments}
The methodology discussed in this paper has been supported by the INRA-1077-SE grant from the French Institute for Agricultural Research (agriculture, food \& nutrition, environment and basic biology). Special thanks to Dr. Isabelle Hue, at INRA Jouy-en-Josas, specialist of molecular biology and embryology and Dr. Christian Meyer, at INRA Versailles, specialist of molecular biology and plant development.

\nocite{*}
\bibliographystyle{plain}
\bibliography{TurenneSNN}
\end{document}